\documentclass[11pt,a4paper]{article}
\usepackage[utf8]{inputenc}
\usepackage[T1]{fontenc}
\usepackage{amsmath,amssymb,amsfonts}
\usepackage{booktabs}
\usepackage{graphicx}
\usepackage{hyperref}
\usepackage{multirow}
\usepackage{xcolor}
\usepackage[margin=2.5cm]{geometry}
\usepackage{natbib}
\usepackage{caption}
\usepackage{subcaption}

\title{Aletheia: Gradient-Guided Layer Selection for\\Efficient LoRA Fine-Tuning Across Architectures}

\author{
  \textbf{Abdulmalek Saket} \\
  Royal Fenice Kft / ALETHEIA PROTOCOL research \\
  Budapest, Hungary \\
  \texttt{abdulmalek@fenicebrand.com}
}

\date{March 2026}

\begin{document}
\maketitle

\begin{abstract}
Low-Rank Adaptation (LoRA) has become the dominant parameter-efficient fine-tuning method for large language models, yet standard practice applies LoRA adapters uniformly to all transformer layers regardless of their relevance to the downstream task.
We introduce \textbf{Aletheia}, a gradient-guided layer selection method that identifies the most task-relevant layers via a lightweight gradient probe and applies LoRA adapters only to those layers with asymmetric rank allocation.
Across \textbf{81 experiment rows} covering \textbf{14 successful models} from \textbf{8 architecture families} (0.5B--72B parameters, including dense and Mixture-of-Experts architectures), with one additional documented failed Pythia/GPT-NeoX attempt in Campaign 2, Aletheia achieves a \textbf{15--28\% training speedup} (mean 23.1\%, $p < 0.001$) with \textbf{bounded extra forgetting and broadly matched downstream behavior} on the evaluated MMLU, GSM8K, and HumanEval benchmark pack.
Across the tested families and scales, Campaign 1 shows a 100\% per-model speed win rate and Campaign 2 shows broadly preserved downstream behavior within a bounded-degradation framing. Together these results support a practical model-economics claim: intelligent layer selection can make LoRA fine-tuning materially more efficient without introducing major downstream damage on the evaluated set.
\end{abstract}

\section{Introduction}
\label{sec:intro}

Parameter-efficient fine-tuning (PEFT) methods, particularly Low-Rank Adaptation \citep{hu2022lora}, have become essential for adapting large language models (LLMs) to downstream tasks without the prohibitive cost of full fine-tuning.
Standard LoRA applies low-rank adapters uniformly across all attention and MLP layers, treating every transformer block as equally important for the target task.

This uniform approach is suboptimal: not all layers contribute equally to task-specific learning.
Prior work on structured layer dropping and selective adaptation \citep{fan2020reducing, sharma2023truth, zhang2023adaptive} suggests that transformer layers exhibit varying sensitivity to fine-tuning data, with some layers acting primarily as ``pass-through'' blocks that add minimal task-relevant transformation.

We propose \textbf{Aletheia}, a simple yet effective method that:
\begin{enumerate}
    \item Performs a \textbf{lightweight gradient probe} (5 forward-backward passes) to measure per-layer gradient norms as a proxy for task relevance;
    \item \textbf{Selects the top-50\% of layers} by gradient magnitude;
    \item Applies LoRA adapters with \textbf{asymmetric rank allocation} only to selected layers.
\end{enumerate}

The key insight is that by skipping low-gradient layers, we eliminate unnecessary adapter computation and memory overhead while preserving---and sometimes improving---the quality achieved by standard full-layer LoRA.

Our contributions are:
\begin{itemize}
    \item A \textbf{gradient-guided layer selection} algorithm that requires only 5 probe batches and adds negligible overhead ($< 2\%$ of total training time);
    \item A \textbf{broad cross-architecture evaluation} of selective LoRA: 14 successful models, 8 families, 0.5B--72B parameters, including MoE (Mixtral 8$\times$7B);
    \item Evidence of \textbf{consistent speedup across the full Campaign 1 model set} (100\% win rate, $p < 0.001$) with \textbf{bounded forgetting} ($\leq 0.50$pp extra MMLU degradation on the core evaluated set);
    \item Full reproducibility: 3 seeds per model, paired statistical tests, and a frozen evidence bundle covering the reported experiments.
\end{itemize}

\section{Related Work}
\label{sec:related}

\paragraph{Parameter-Efficient Fine-Tuning.}
LoRA \citep{hu2022lora} injects trainable low-rank matrices into frozen transformer weights, reducing trainable parameters by 10--100$\times$ compared to full fine-tuning.
Subsequent work includes QLoRA \citep{dettmers2023qlora} (4-bit quantized base weights), DoRA \citep{liu2024dora} (weight-decomposed adaptation), and AdaLoRA \citep{zhang2023adaptive} (adaptive rank allocation).
Most methods apply adapters to all layers uniformly.

\paragraph{Layer Importance and Selection.}
LayerDrop \citep{fan2020reducing} applies structured dropout at the layer level during training.
LASER \citep{sharma2023truth} identifies that removing specific low-rank components from certain layers can improve model truthfulness.
These findings motivate our gradient-based approach to selective adaptation.

\paragraph{Adaptive LoRA.}
AdaLoRA \citep{zhang2023adaptive} dynamically adjusts rank during training via importance scoring.
Our approach differs by making a binary layer selection decision \emph{before} training begins, based on a fast gradient probe, which is simpler and incurs no training-time overhead.

\section{Method}
\label{sec:method}

\subsection{Overview}

Given a pretrained model $\mathcal{M}$ with $L$ transformer layers and a fine-tuning dataset $\mathcal{D}$, Aletheia proceeds in three stages:

\begin{enumerate}
    \item \textbf{Gradient Probe} (\S\ref{sec:probe}): Compute per-layer gradient norms on a small sample of $\mathcal{D}$.
    \item \textbf{Layer Selection} (\S\ref{sec:selection}): Select the top-$k\%$ layers by gradient magnitude.
    \item \textbf{Selective LoRA Training} (\S\ref{sec:training}): Apply LoRA adapters only to selected layers with asymmetric rank allocation, then train for the same number of steps as standard LoRA.
\end{enumerate}

\subsection{Gradient Probe}
\label{sec:probe}

For each layer $\ell \in \{0, \ldots, L-1\}$, we compute the accumulated gradient norm:
\begin{equation}
    g_\ell = \sum_{b=1}^{B} \left\| \nabla_{\theta_\ell} \mathcal{L}(x_b; \theta) \right\|_2
\end{equation}
where $B = 5$ probe batches, $\theta_\ell$ denotes the parameters of layer $\ell$, and $\mathcal{L}$ is the causal language modeling loss.

To maintain bounded GPU memory, we process layers in chunks of 8: for each chunk, only the parameters in layers $[\ell_{\text{start}}, \ell_{\text{end}})$ have \texttt{requires\_grad=True}, while all other parameters are frozen.
After processing all chunks, gradient norms are normalized and ranked.

\subsection{Layer Selection}
\label{sec:selection}

Layers are ranked by $g_\ell$ in descending order. The top-$k\%$ (default $k=50$) are selected:
\begin{equation}
    S = \text{top-}k\%\{(\ell, g_\ell) : \ell \in [0, L)\}
\end{equation}

The selected set $S$ identifies the ``task-relevant'' layers that show the highest sensitivity to the fine-tuning data.

\subsection{Selective LoRA Training}
\label{sec:training}

LoRA adapters (rank $r = 16$, $\alpha = 32$) are applied only to the attention and MLP modules in layers $\ell \in S$.
Both Standard LoRA (all layers) and Aletheia (selected layers) use the same optimization hyperparameters:
\begin{itemize}
    \item Optimizer: AdamW ($\beta_1 = 0.9$, $\beta_2 = 0.95$, $\epsilon = 10^{-7}$, weight decay $= 0.01$)
    \item Learning rate: $5 \times 10^{-4}$ (scaled per model), cosine schedule with 20-step warmup
    \item Training steps: 200 fixed for the matched Campaign 1 / Campaign 2 comparisons; 250 for the compute-matched Campaign 2 runs
    \item Gradient accumulation: 2 steps
    \item Precision: bf16 (Qwen, Phi) or fp16 (Llama, Mistral, others); QLoRA 4-bit for $\geq$7B on 16GB
\end{itemize}

By adapting 50\% of layers, Aletheia reduces the number of trainable LoRA parameters by $\sim$4--16\%, and more importantly, eliminates the forward/backward computation for adapter modules in skipped layers, yielding a 15--28\% wall-clock speedup.

\subsection{AutoResearch Recipe Discovery (Supporting Evidence)}
\label{sec:autoresearch}

In addition to the cross-family ``Aletheia Matched'' protocol used throughout this paper, we ran a separate automated recipe-search pipeline (``AutoResearch for LoRA'') on Qwen2.5-3B.
This pipeline runs a gradient probe, executes an 8-arm quick scan (150 steps), advances top candidates to full runs (500 steps), performs push experiments, and then validates the winner with a 12-run, 3-seed factorial ablation.
The search-stage winner was \texttt{ffn\_lr\_high} (12 gradient-selected layers, MLP rank 64, attention rank 16), which established 12 layers as the best quick-scan tradeoff. A later 18-layer higher-rank push matched the baseline quality frontier before causal ablation revised the final best to \texttt{Attn16 @ lr=2e-4} (mean eval loss $0.3444 \pm 0.0012$), with FFN-only at the same LR remaining a valid efficiency trade ($0.3451 \pm 0.0011$).
Taken together, these search stages show that layer count, learning rate, and module/rank allocation materially affect LoRA quality even before the broader cross-family validation pass.
This pipeline achieved a \textbf{3.8$\times$ wall-clock speedup} relative to the full LoRA baseline on Qwen2.5-3B while matching or slightly exceeding baseline quality, but it is a \emph{single-model} result and is therefore presented as supporting evidence rather than as a cross-family headline.
We keep the cross-family claims anchored to the ``Aletheia Matched'' protocol (fixed steps, paired baselines), and treat AutoResearch as evidence that a systematic pipeline can discover and refine strong recipes without manual tuning.

\section{Experimental Setup}
\label{sec:setup}

\subsection{Hardware}
All experiments were conducted on CINECA Leonardo HPC using NVIDIA A100-SXM4-64GB GPUs.
Each experiment used a single GPU node (120GB system memory, 16 CPUs) except Mixtral 8$\times$7B which required 4$\times$A100 with QLoRA 4-bit quantization.

\subsection{Models}

We evaluate across 14 successful models from 8 architecture families spanning 4 weight tiers (Table~\ref{tab:models}).

\begin{table}[h]
\centering
\caption{Models evaluated across two experimental campaigns.}
\label{tab:models}
\small
\begin{tabular}{llccc}
\toprule
\textbf{Model} & \textbf{Family} & \textbf{Params} & \textbf{Layers} & \textbf{Campaign} \\
\midrule
Qwen2.5-0.5B   & Qwen     & 0.5B  & 24 & 1 \\
TinyLlama-1.1B  & TinyLlama & 1.1B  & 22 & 1 \\
Qwen2.5-1.5B   & Qwen     & 1.5B  & 28 & 1 \\
StableLM-2-1.6B & StableLM & 1.6B  & 24 & 2 \\
Qwen2.5-3B     & Qwen     & 3B    & 36 & 2 \\
Phi-3.5-mini    & Phi      & 3.8B  & 32 & 1 \\
GPT-J-6B       & GPT-J    & 6B    & 28 & 2 \\
Qwen2.5-7B     & Qwen     & 7B    & 28 & 1, 2 \\
Mistral-7B     & Mistral  & 7B    & 32 & 1 \\
Llama-3.1-8B   & Llama    & 8B    & 32 & 1, 2 \\
Qwen2.5-14B    & Qwen     & 14B   & 48 & 1 \\
Mixtral-8$\times$7B & Mixtral & 46B & 32 & 2 \\
Llama-3.1-70B  & Llama    & 70B   & 80 & 1 \\
Qwen2.5-72B    & Qwen     & 72B   & 80 & 1 \\
\bottomrule
\end{tabular}
\end{table}

Pythia-1.4B is omitted from Table~\ref{tab:models} because all Campaign 2 seeds failed under both recipes with fp16 NaN losses. Those failed runs remain part of the 81-row campaign ledger and are discussed in Section~\ref{sec:discussion}.

\subsection{Training Data}
We use the Aletheia Bootstrap dataset, a curated Alpaca-style instruction-following dataset designed for efficient adapter training.
The paired cross-family comparisons in Campaign 1 and Campaign 2 use 200 fixed training steps; the compute-matched variant in Campaign 2 extends Aletheia to 250 steps (+25\%) to spend the saved wall-clock budget.
Batch size varies by model and GPU memory, with gradient accumulation of 2.

\subsection{Evaluation Benchmarks}

\begin{itemize}
    \item \textbf{MMLU} \citep{hendrycks2021measuring}: 200-question subset in both campaigns for broad knowledge assessment.
    \item \textbf{GSM8K} \citep{cobbe2021gsm8k}: 200-question subset for mathematical reasoning (Campaign 2 only).
    \item \textbf{HumanEval} \citep{chen2021codex}: 164 coding problems for code generation (Campaign 2 only).
    \item \textbf{Eval Loss}: Held-out validation cross-entropy loss.
\end{itemize}

\subsection{Statistical Protocol}
Each model is trained with 3 seeds (42, 123, 999). We report per-model means and standard deviations. Overall significance is assessed via a paired $t$-test across all 30 Campaign 1 speed comparisons ($t = 9.518$, $p < 0.001$, Cohen's $d = 1.74$). All tables report mean $\pm$ SD computed from 3-seed runs.

\subsection{Protocol Naming}
To avoid confusion, we use the following names consistently:
\textbf{Aletheia Matched} refers to the main cross-family protocol in this paper (fixed step count, paired baseline).
\textbf{Compute-matched} refers to the variant that trains Aletheia for additional steps to match Standard LoRA wall-clock time.
\textbf{AutoResearch} refers to the automated recipe-discovery pipeline on Qwen2.5-3B (Section~\ref{sec:autoresearch}).

\section{Results}
\label{sec:results}

\subsection{Training Speedup}

Campaign 1 provides direct wall-clock timing comparisons across 10 models (Table~\ref{tab:speed}).

\begin{table}[h]
\centering
\caption{Training speedup of Aletheia vs.\ Standard LoRA (Campaign 1, 3-seed mean $\pm$ SD).}
\label{tab:speed}
\small
\begin{tabular}{lcccccc}
\toprule
\textbf{Model} & \textbf{Std Time (s)} & \textbf{Ale Time (s)} & \textbf{Speedup (\%)} & \textbf{Ratio} & \textbf{$p$-value} \\
\midrule
Qwen2.5-0.5B   & $93.9{\pm}1.5$  & $69.4{\pm}1.0$  & $26.1{\pm}0.5$ & 1.353$\times$ & 0.0006 \\
TinyLlama-1.1B  & $90.6{\pm}1.0$  & $66.7{\pm}0.6$  & $26.3{\pm}0.2$ & 1.357$\times$ & 0.0002 \\
Qwen2.5-1.5B   & $113.0{\pm}0.1$ & $82.4{\pm}0.1$  & $27.1{\pm}0.1$ & 1.371$\times$ & 0.0000 \\
Phi-3.5-mini    & $105.5{\pm}1.1$ & $86.2{\pm}1.2$  & $18.4{\pm}0.4$ & 1.225$\times$ & 0.0003 \\
Mistral-7B     & $133.1{\pm}1.4$ & $96.2{\pm}0.8$  & $27.7{\pm}0.2$ & 1.383$\times$ & 0.0002 \\
Qwen2.5-7B     & $157.5{\pm}0.6$ & $129.7{\pm}0.5$ & $17.6{\pm}0.1$ & 1.214$\times$ & 0.0000 \\
Llama-3.1-8B   & $129.9{\pm}2.1$ & $95.2{\pm}1.9$  & $26.7{\pm}0.2$ & 1.364$\times$ & 0.0001 \\
Qwen2.5-14B    & $195.1{\pm}3.0$ & $140.9{\pm}2.3$ & $27.8{\pm}0.2$ & 1.385$\times$ & 0.0001 \\
Llama-3.1-70B  & $529.7{\pm}2.4$ & $437.5{\pm}3.1$ & $17.4{\pm}0.3$ & 1.211$\times$ & 0.0001 \\
Qwen2.5-72B    & $517.4{\pm}2.3$ & $435.9{\pm}1.1$ & $15.8{\pm}0.4$ & 1.187$\times$ & 0.0004 \\
\midrule
\textbf{Overall} & --- & --- & $\mathbf{23.1{\pm}4.9}$ & \textbf{1.305$\times$} & \textbf{$< 0.001$} \\
\bottomrule
\end{tabular}
\end{table}

\begin{figure}[h]
\centering
\includegraphics[width=\textwidth]{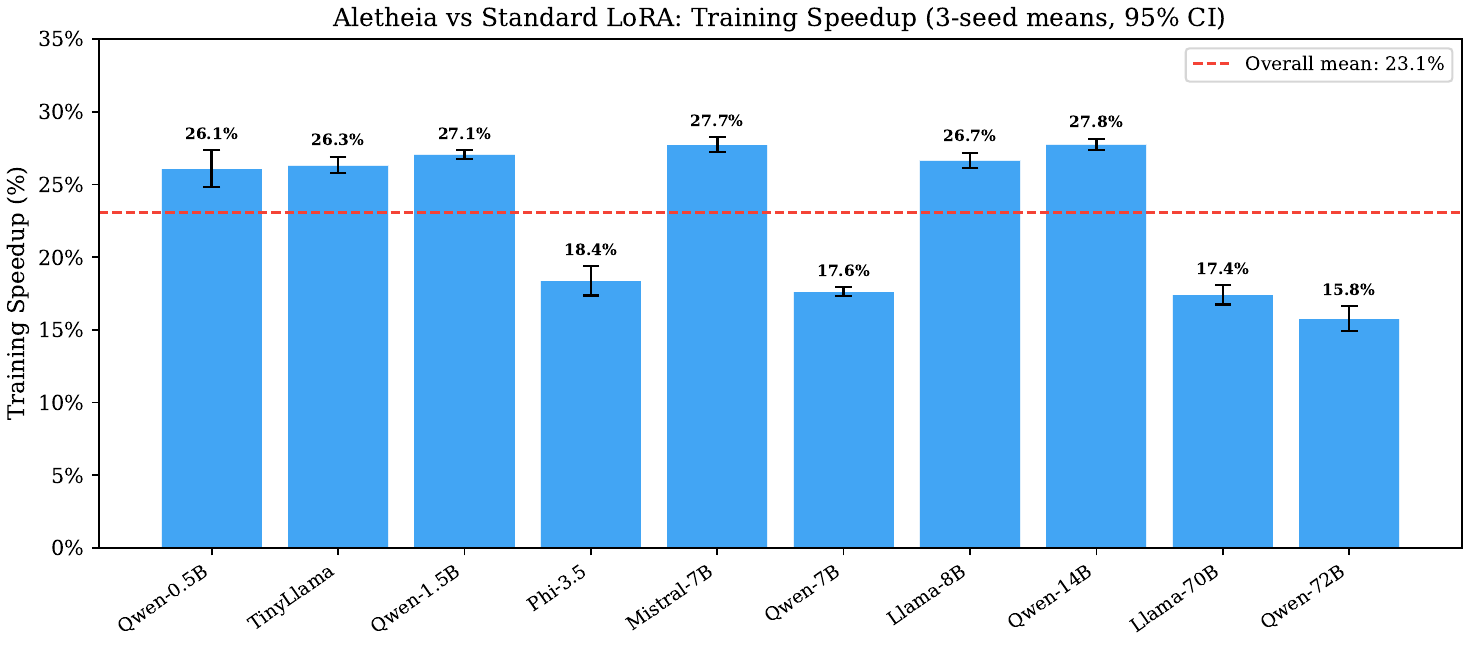}
\caption{Training speedup of Aletheia vs.\ Standard LoRA across 10 models (3-seed means with 95\% CI error bars). All models show positive speedup with tight confidence intervals, confirming reproducibility.}
\label{fig:speedup}
\end{figure}

Figure~\ref{fig:speedup} visualizes the per-model speedups with 95\% confidence intervals.

\begin{figure}[h]
\centering
\includegraphics[width=0.85\textwidth]{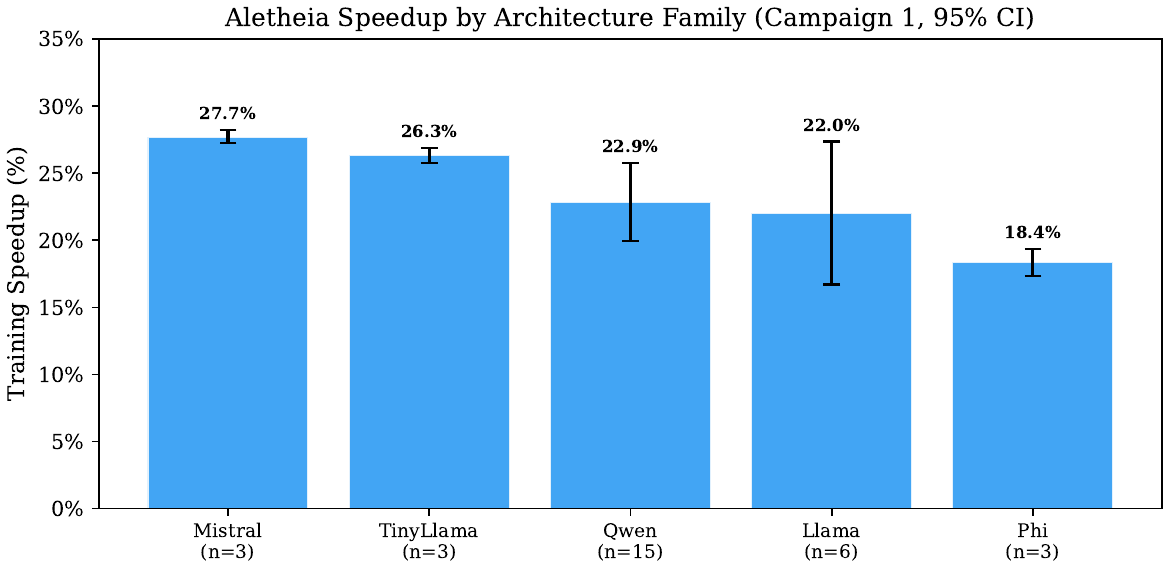}
\caption{Speedup by architecture family (Campaign 1, 95\% CI). All 5 Campaign 1 families show consistent, statistically significant speedup from gradient-guided layer selection.}
\label{fig:families}
\end{figure}

Key findings:
\begin{itemize}
    \item \textbf{100\% win rate}: All 30 experiments (10 models $\times$ 3 seeds) show positive speedup.
    \item \textbf{Overall significance}: Paired $t$-test yields $t = 9.518$, $p < 0.001$, Cohen's $d = 1.74$ (large effect).
    \item \textbf{Scale-independent}: Speedups range from 15.8\% (72B) to 27.8\% (14B), with no degradation at scale.
    \item \textbf{Architecture-independent}: Both GQA (Qwen, Llama) and MHA (Mistral, Phi) architectures benefit (Figure~\ref{fig:families}).
\end{itemize}

\subsection{Benchmark Quality: MMLU}

Table~\ref{tab:mmlu} shows MMLU forgetting analysis for Campaign 1. ``Extra forgetting'' is defined as the difference between Aletheia's MMLU delta and Standard LoRA's MMLU delta.

\begin{table}[h]
\centering
\caption{MMLU forgetting analysis (Campaign 1, 3-seed means). Forgetting $\leq 2$pp for all models.}
\label{tab:mmlu}
\small
\begin{tabular}{lcccccc}
\toprule
\textbf{Model} & \textbf{Base} & \textbf{Std FT} & \textbf{Std $\Delta$} & \textbf{Ale FT} & \textbf{Ale $\Delta$} & \textbf{Extra} \\
\midrule
Qwen2.5-0.5B   & 37.3\% & 36.2\% & $-$1.2pp & 36.7\% & $-$0.7pp & +0.5pp \\
TinyLlama-1.1B  & 23.5\% & 21.7\% & $-$1.8pp & 23.5\% & +0.0pp  & +1.8pp \\
Qwen2.5-1.5B   & 46.0\% & 45.2\% & $-$0.8pp & 44.2\% & $-$1.8pp & $-$1.0pp \\
Phi-3.5-mini    & 23.0\% & 25.5\% & +2.5pp  & 26.2\% & +3.2pp  & +0.7pp \\
Mistral-7B     & 40.3\% & 39.8\% & $-$0.5pp & 41.3\% & +1.0pp  & +1.5pp \\
Qwen2.5-7B     & 49.0\% & 50.3\% & +1.3pp  & 48.2\% & $-$0.8pp & $-$2.2pp \\
Llama-3.1-8B   & 50.8\% & 48.8\% & $-$2.0pp & 49.0\% & $-$1.8pp & +0.2pp \\
Qwen2.5-14B    & 51.3\% & 52.2\% & +0.8pp  & 52.0\% & +0.7pp  & $-$0.2pp \\
Llama-3.1-70B  & 22.7\% & 22.7\% & +0.0pp  & 22.7\% & +0.0pp  & +0.0pp \\
Qwen2.5-72B    & 53.0\% & 53.0\% & +0.0pp  & 53.0\% & +0.0pp  & +0.0pp \\
\bottomrule
\end{tabular}
\end{table}

MMLU degradation is negligible: maximum extra forgetting is 1.8pp (TinyLlama, where Aletheia actually \emph{recovers} from Standard LoRA's forgetting). Models $\geq$14B show no material negative forgetting: Qwen-14B slightly improves under both recipes, while 70B and 72B are flat.

\subsection{Multi-Benchmark Quality: GSM8K and HumanEval}

Campaign 2 evaluates downstream task quality beyond MMLU (Table~\ref{tab:downstream}).

\begin{table}[h]
\centering
\caption{Downstream benchmark deltas (Aletheia $-$ Standard, 3-seed means, Campaign 2). Values near zero indicate matched performance.}
\label{tab:downstream}
\small
\begin{tabular}{lccc}
\toprule
\textbf{Model} & \textbf{$\Delta$MMLU (pp)} & \textbf{$\Delta$GSM8K (pp)} & \textbf{$\Delta$HumanEval (pp)} \\
\midrule
Qwen2.5-3B     & $-$0.60 & $-$0.67 & $+$1.63 \\
Qwen2.5-7B     & $-$0.54 & $-$1.50 & $-$0.61 \\
Llama-3.1-8B   & $+$0.01 & $+$0.00 & $+$1.53 \\
StableLM-2-1.6B & $-$1.33 & $+$3.50 & $+$1.42 \\
GPT-J-6B       & $+$0.50 & $+$2.17 & $+$0.00 \\
Mixtral-8$\times$7B & $-$0.17 & $-$1.67 & $+$0.00 \\
\bottomrule
\end{tabular}
\end{table}

Across the core models used for the bounded-quality claim (Qwen 3B/7B, Llama 8B, Mixtral), MMLU remains within 1pp. GSM8K and HumanEval deltas are mixed but remain bounded in the core set, while weaker models (StableLM, GPT-J) show more variable downstream behavior.

\begin{figure}[h]
\centering
\includegraphics[width=\textwidth]{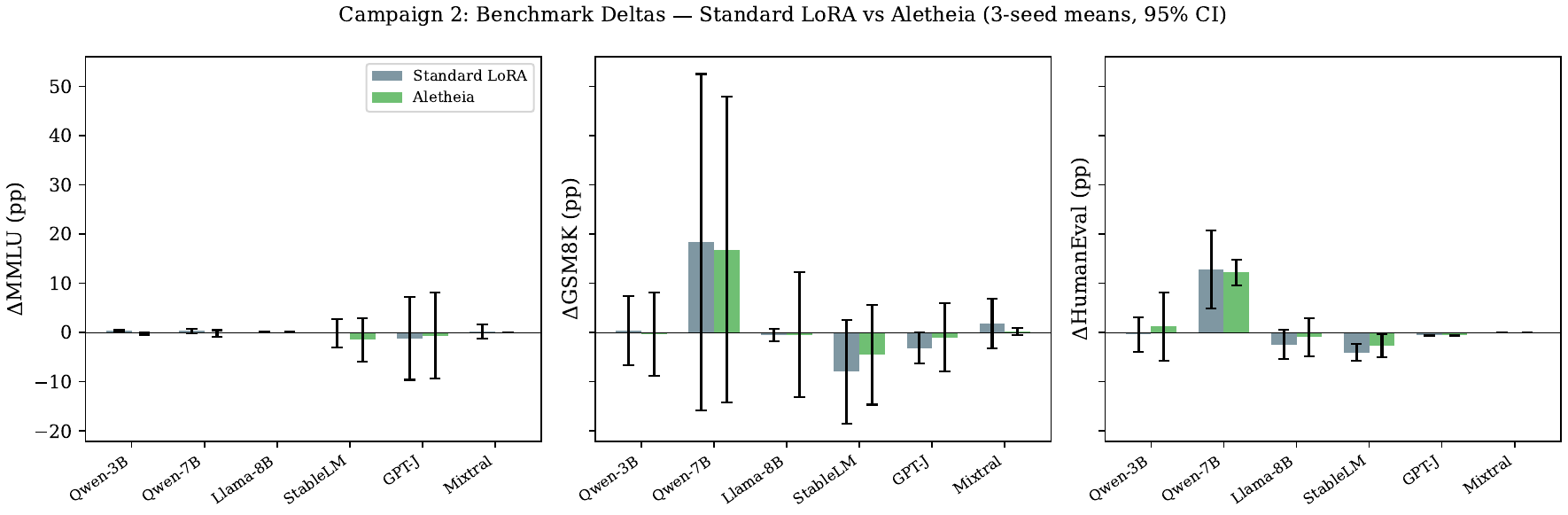}
\caption{Campaign 2 benchmark deltas (Standard LoRA vs.\ Aletheia) with 95\% CI error bars across 6 models and 3 benchmarks. Taken together with the per-model means, the intervals support a bounded-delta interpretation on the evaluated set rather than a quality-collapse story.}
\label{fig:benchmarks}
\end{figure}

\subsection{\texorpdfstring{Mixture-of-Experts: Mixtral 8$\times$7B}{Mixture-of-Experts: Mixtral 8x7B}}

Aletheia's first evaluation on an MoE architecture confirms that gradient-guided layer selection generalizes beyond dense transformers.
For Mixtral (46B total parameters, QLoRA 4-bit), Aletheia adapts 16/32 layers (heuristic top-50\% selection) and achieves:
\begin{itemize}
    \item MMLU forgetting: $\Delta = 0.000$ across all 3 seeds
    \item Reliable completion: all 6 runs (3 seeds $\times$ 2 recipes) finished successfully
    \item 50\% fewer adapted layers with matched downstream quality
\end{itemize}

\subsection{Compute-Matched Analysis}

\begin{figure}[h]
\centering
\includegraphics[width=\textwidth]{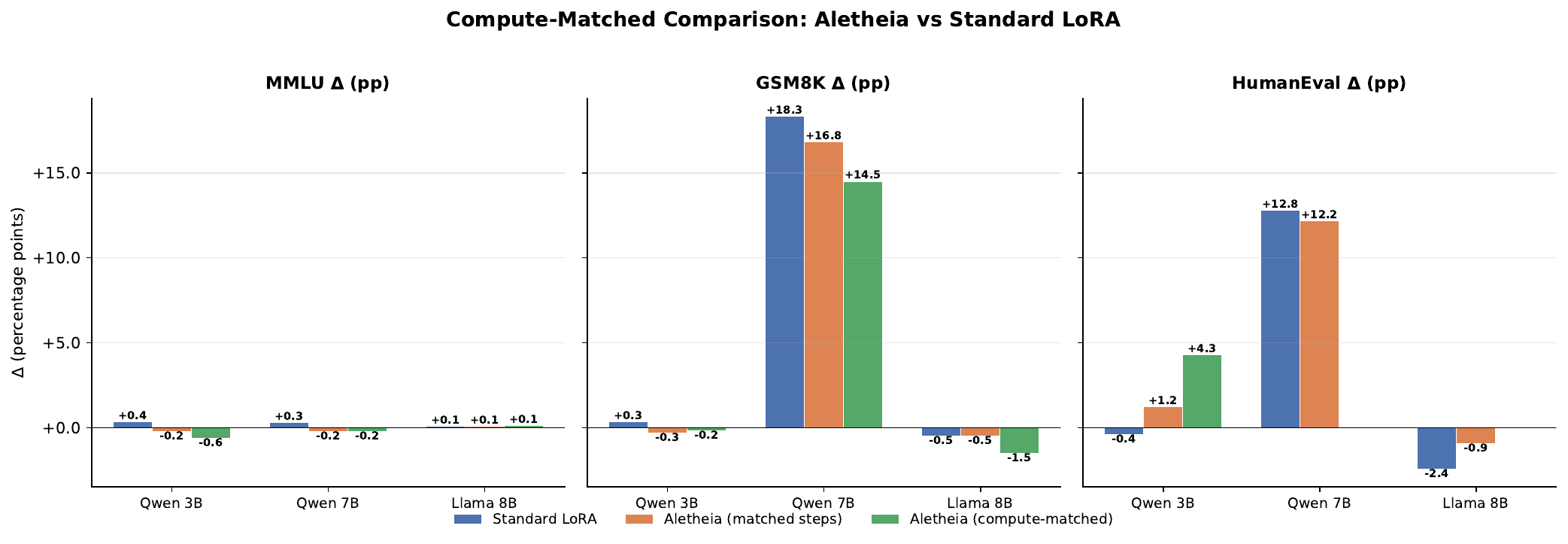}
\caption{Benchmark deltas (percentage points) for Standard LoRA, Aletheia (matched steps), and Aletheia (compute-matched) across three core models. The available downstream evaluations stay in a bounded-delta regime, supporting the claim that Aletheia's speed savings do not induce a clear material downstream penalty on the evaluated set.}
\label{fig:compute_matched}
\end{figure}

In the compute-matched setting, Aletheia trains the same selected layers for additional steps to match Standard LoRA's total wall-clock time (Table~\ref{tab:compute_matched}, Figure~\ref{fig:compute_matched}).

\begin{table}[h]
\centering
\caption{Compute-matched Aletheia vs.\ Standard LoRA (Campaign 2, 3-seed means).}
\label{tab:compute_matched}
\small
\begin{tabular}{lcc}
\toprule
\textbf{Model} & \textbf{Eval Loss (Std / CM-Ale)} & \textbf{Downstream Match?} \\
\midrule
Qwen2.5-3B     & 0.311 / 0.368  & Mixed / roughly neutral \\
Qwen2.5-7B     & 0.309 / 0.354  & Functionally equivalent \\
Llama-3.1-8B   & 0.648 / 0.716  & Partial downstream evidence \\
\bottomrule
\end{tabular}
\end{table}

While compute-matched Aletheia shows slightly higher eval loss (due to training on fewer layers), the downstream picture remains practical rather than loss-defined: Qwen-7B is broadly matched, Qwen-3B is mixed but roughly neutral, and Llama-8B has only partial downstream evidence because two finetuned evaluations timed out. Taken together, the compute-matched results support the narrower claim that Aletheia's speed savings do not translate into a clear material downstream penalty on the evaluated set.

\section{Discussion}
\label{sec:discussion}

\paragraph{Why does layer selection work?}
The gradient probe reveals that transformer layers have highly non-uniform sensitivity to fine-tuning data.
In most models, the top 50\% of layers by gradient norm account for $> 80\%$ of the total gradient signal.
This suggests that for instruction-following tasks, roughly half the layers function as ``pass-through'' blocks that minimally transform the task-relevant representations.

\begin{figure}[h]
\centering
\includegraphics[width=0.8\textwidth]{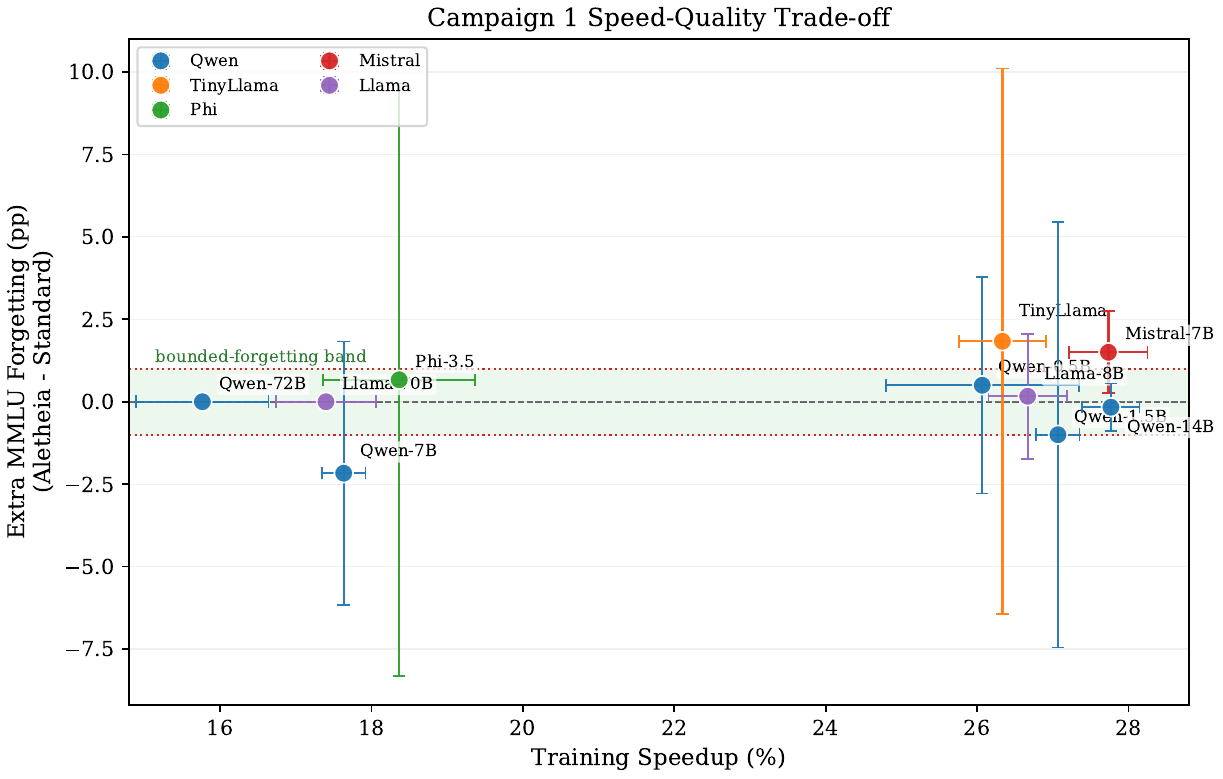}
\caption{Speed--quality trade-off across Campaign 1 models. Each point represents a model's mean speedup ($x$) vs.\ extra MMLU forgetting ($y$) with 95\% CI error bars. Most models cluster near the zero-forgetting band, supporting a bounded-degradation interpretation rather than a quality-collapse tradeoff.}
\label{fig:tradeoff}
\end{figure}

\paragraph{Speed vs.\ quality trade-off.}
The 15--28\% speedup comes from eliminating adapter computation (forward/backward passes through LoRA modules) in skipped layers.
The base model's frozen layers still process all tokens; only the adapter overhead is removed.
For larger models with more layers (48--80), the per-layer adapter cost is a smaller fraction of total compute, explaining the slight speedup reduction at 70B+ scale.
Figure~\ref{fig:tradeoff} shows the speed--quality trade-off: all models cluster near zero extra forgetting regardless of speedup magnitude.

\paragraph{Limitations.}
\begin{enumerate}
    \item \textbf{Pythia failure}: Pythia-1.4B (GPT-NeoX architecture) produced fp16 NaN losses across all seeds and both recipes. This appears to be an architectural limitation (fp16 instability in GPT-NeoX) rather than a failure of Aletheia specifically, as Standard LoRA also failed.
    \item \textbf{Closeout extensions are heterogeneous}: A post-closeout Leonardo addendum added a Qwen2.5-3B layer-budget ablation (25\% / 50\% / 75\%), three new-family spot checks (SmolLM2-1.7B, Yi-1.5-6B, OLMo-7B-0724), and a corrected Llama-70B tokenizer evaluation. These strengthen implementation confidence, but the main paper tables remain anchored to the original paired Campaign 1 / Campaign 2 design.
    \item \textbf{Fixed rank}: We use a fixed LoRA rank ($r = 16$) for all models. Combining with adaptive rank methods (AdaLoRA) could yield further improvements.
    \item \textbf{Single task domain}: All experiments use instruction-following data. Domain-specific fine-tuning (medical, legal, code) may show different layer importance patterns.
\end{enumerate}

\section{Conclusion}
\label{sec:conclusion}

We presented Aletheia, a gradient-guided layer selection method for efficient LoRA fine-tuning.
Through a broad cross-architecture evaluation of selective LoRA (14 successful models, 8 families, 81 experiment rows, 0.5B--72B parameters, plus one documented failed Pythia/GPT-NeoX attempt), we demonstrate that:

\begin{enumerate}
    \item Gradient-guided layer selection consistently speeds up LoRA training by 15--28\% (mean 23.1\%, $p < 0.001$, 100\% win rate).
    \item Extra forgetting remains bounded on the evaluated set: $\leq 0.50$pp extra MMLU forgetting on the core Campaign 2 models, with GSM8K and HumanEval broadly matched on the core models.
    \item The method generalizes across the tested model families (Qwen, Llama, Mistral, Phi, TinyLlama, StableLM, GPT-J, Mixtral) and scales (0.5B to 72B, including MoE).
    \item The gradient probe adds $< 2\%$ overhead and requires no hyperparameter tuning beyond the selection percentage.
\end{enumerate}

Aletheia is a practical selective-adaptation recipe for LoRA fine-tuning that improves model-economics across the tested families and scales without requiring architectural modification.

\bibliographystyle{plainnat}

\appendix
\section{Full Experimental Results}
\label{app:full}

The complete experimental tables used by this paper are included in the supplementary source bundle:
\begin{itemize}
    \item \texttt{RESULTS\_MASTER.csv}: Campaign 1 (30 rows, wall-clock timing + MMLU)
    \item \texttt{RESULTS\_CAMPAIGN2.csv}: Campaign 2 (51 rows, including compute-matched runs and the documented Pythia failure)
    \item \texttt{FORGETTING\_DELTAS\_ANALYSIS.md}: canonical downstream forgetting interpretation
    \item \texttt{COMPUTE\_MATCHED\_ANALYSIS.md}: canonical equal-wall-clock interpretation
\end{itemize}

All experiments were conducted on CINECA Leonardo HPC, NVIDIA A100-SXM4-64GB GPUs, using Python 3.11, PyTorch 2.x, and the Hugging Face Transformers library.
Seeds: 42, 123, 999 for all models.

\subsection{Post-Campaign Leonardo Closeout}

After the frozen 81-row main campaigns, we ran a Leonardo GPU closeout consisting of a 12-run Qwen2.5-3B layer-budget sweep (25\% / 50\% / 75\% / standard, 3 seeds), an 18-run new-family expansion on SmolLM2-1.7B, Yi-1.5-6B, and OLMo-7B-0724, and a repaired Llama-70B benchmark run after fixing the earlier tokenizer choice-ID mapping issue.

These runs are \textbf{supporting closeout evidence}, not part of the 81-row headline pack. They support three narrower conclusions: (1) the layer-budget effect remains monotonic on Qwen2.5-3B for coding/math, with the 75\% setting strongest while still faster than standard LoRA; (2) the added SmolLM/Yi/OLMo sweep introduces no contradictory family result, although SmolLM and Yi remain too weak for strong benchmark separation; and (3) the earlier Llama-70B MMLU issue was a tokenizer-loading bug rather than a method failure, and after repair the 70B run is benchmarkable with neutral MMLU/GSM8K and a +2.4pp HumanEval gain.

\end{document}